\pdfoutput=1
\documentclass[11pt]{article}

\usepackage{acl}

\usepackage{times}
\usepackage{latexsym}

\usepackage[T1]{fontenc}
\usepackage[utf8]{inputenc}

\usepackage{microtype}

\usepackage{amsmath}
\usepackage{booktabs}
\usepackage{multirow}
\usepackage{graphicx}
\usepackage{enumitem}
\usepackage{amssymb}
\usepackage{xcolor}
\usepackage{colortbl}

\definecolor{Gray}{gray}{0.9}
\definecolor{LightCyan}{rgb}{0.88,1,1}

\newcolumntype{a}{>{\columncolor{Gray}}c}
\newcolumntype{b}{>{\columncolor{white}}c}

\newcommand{\bleu}{{\small BLEU-4}}
\newcommand{\rouge}{{\small ROUGE-L}}
\newcommand{\meteor}{{\small METEOR}}
\newcommand{\cider}{{\small CIDEr}}
\newcommand{\spice}{{\small SPICE}}
\newcommand{\infometric}{InfoMetIC}

\title{InfoMetIC: An Informative Metric for \\ Reference-free Image Caption Evaluation}

\author{Anwen Hu\textsuperscript{\rm 1}, Shizhe Chen\textsuperscript{\rm 2}, LiangZhang\textsuperscript{\rm 1}, Qin Jin\textsuperscript{\rm 1}\\
  \textsuperscript{\rm 1}School of Information, Renmin University of China \\
  \textsuperscript{\rm 2}INRIA \\
  \texttt{\{anwenhu,zhangliang00,qjin\}@ruc.edu.cn} \\
  \texttt{shizhe.chen@inria.fr}}

\begin{document}
\maketitle
\begin{abstract}
Automatic image captioning evaluation is critical for benchmarking and promoting advances in image captioning research. Existing metrics only provide a single score to measure caption qualities, which are less explainable and informative. Instead, we humans can easily identify the problems of captions in details, e.g., which words are inaccurate and which salient objects are not described, and then rate the caption quality. To support such informative feedback, we propose an \textbf{Info}rmative \textbf{Met}ric for Reference-free \textbf{I}mage \textbf{C}aption evaluation (InfoMetIC). Given an image and a caption, InfoMetIC is able to report incorrect words and unmentioned image regions at fine-grained level, and also provide a text precision score, a vision recall score and an overall quality score at coarse-grained level. The coarse-grained score of InfoMetIC achieves significantly better correlation with human judgements than existing metrics on multiple benchmarks. We also construct a token-level evaluation dataset and demonstrate the effectiveness of InfoMetIC in fine-grained evaluation. Our code and datasets are publicly available at \url{https://github.com/HAWLYQ/InfoMetIC}.

\end{abstract}

\section{Introduction}
Image captioning aims to automatically generate natural language sentences to describe image contents. Recently, there are significant breakthroughs in image captioning such as attention-based model architectures \cite{DBLP:conf/cvpr/00010BT0GZ18,pan2020x,hu2020icecap, hu2021question} and vision-and-language pretraining (VLP) \cite{DBLP:conf/aaai/ZhouPZHCG20,xia2021xgpt,li2022blip, xu2021e2e, xu2023mplug}.
However, as groundtruth image descriptions are extremely diverse and subjective, evaluating the image captioning performance remains a considerable challenge.

The most widely used image captioning metrics such as \meteor~\cite{banerjee2005meteor}, \cider~\cite{vedantam2015cider} and \spice~\cite{anderson2016spice} utilize human-written descriptions of images as references and measure similarities between generated captions and references for evaluation.
Such reference-based approaches suffer from two major limitations.
Firstly, these metrics mainly evaluate caption quality by n-gram overlaps which fail to measure genuine semantic similarities. Secondly, references require time-consuming annotations and thus there are only a few annotated captions (typically 5) for each image. The limited number of references cannot fully capture image contents, resulting in incorrect penalties when generated captions describe correct novel things that are not mentioned in the references.

\begin{figure}[!tbp]
    \centering
    \includegraphics[width=0.95\linewidth]{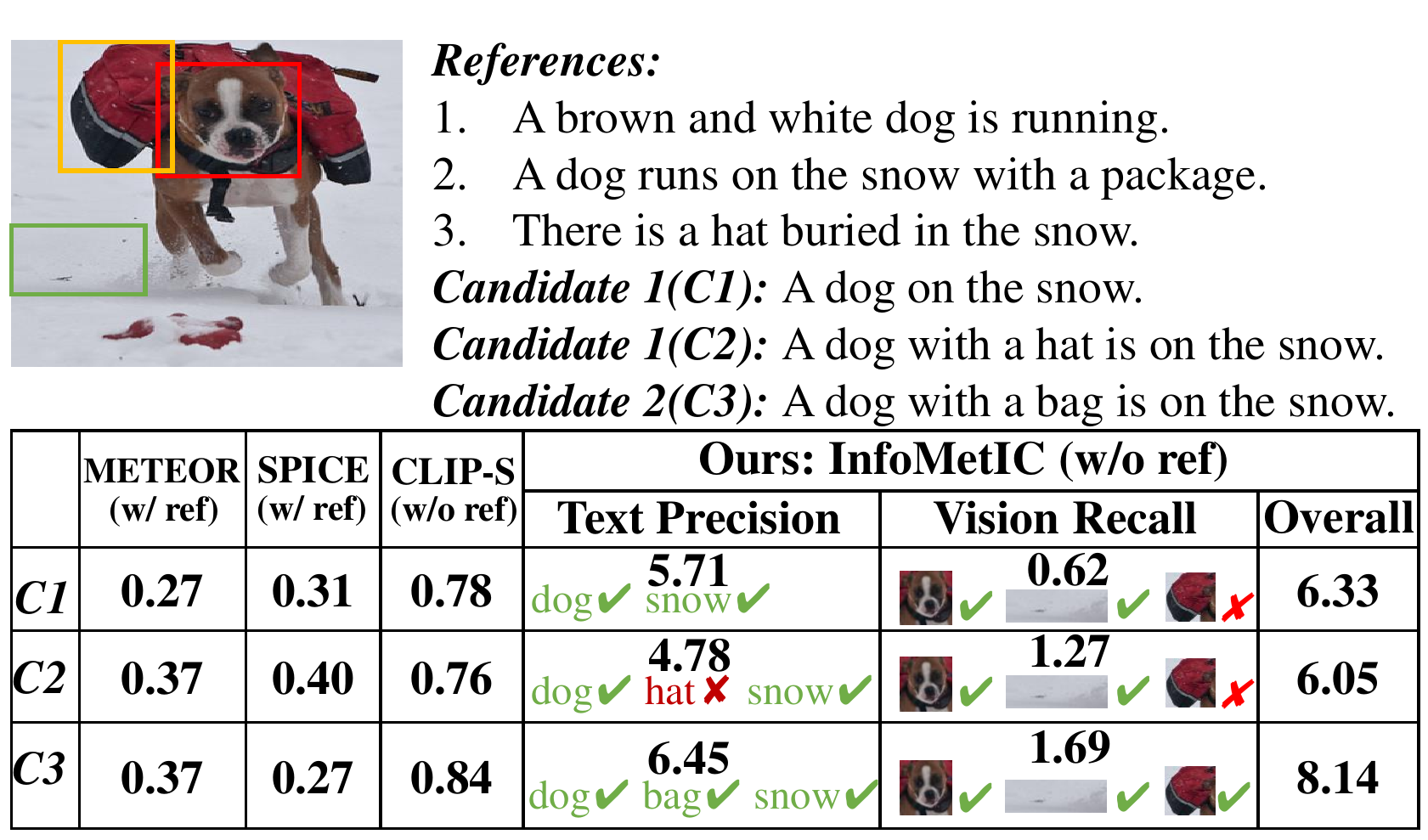}

    \caption{Comparison of existing metrics and our informative metric (InfoMetIC). `w/ ref' and `w/o ref' mean using references or not.}
    \label{fig:intro}
\end{figure}

To alleviate the above limitations, recent works are more focusing on reference-free metrics, which directly use images instead of reference captions in evaluation.
Benefited from the success of VLP on large-scale web data, UMIC \cite{DBLP:conf/acl/Lee0DBJ20} and CLIP-S \cite{DBLP:conf/emnlp/HesselHFBC21} leverage VLP models UNITER \cite{chen2020uniter} and CLIP \cite{DBLP:conf/icml/RadfordKHRGASAM21} respectively to calculate relevance scores between generated captions and images.
Although they have achieved promising correlations with human judgments, they can only produce an overall score as quality measurement.
We humans instead tend to evaluate captions considering two aspects: 1) whether the caption correctly describes the image content (named \emph{text precision}); and 2) whether the image content is comprehensively described in the caption (named \emph{vision recall}).
For example, as shown Figure~\ref{fig:intro}, we can easily tell the ``hat'' in the second candidate is incorrect, and some salient contents such as ``the bag'' are not mentioned, and thus form our final evaluation to the caption.

For the purpose of providing explainable and detailed feedbacks, we propose a \textbf{Info}rmative \textbf{Met}ric for Reference-free \textbf{I}mage \textbf{C}aption evaluation (\infometric). 
It is built on top of pretrained VLP models to measure fine-grained cross-modal similarities.
\infometric~is able to point out incorrect semantic words in the caption and unmentioned regions in the image. Based on fine-grained evaluation, it derives text precision and vision recall scores to measure captioning accuracy and completeness respectively. We take a summation of the two scores to rate overall quality of the caption.

Our contributions in this work are three-fold:
\parskip=0.1em
\begin{itemize}[itemsep=0.1em,parsep=0em,topsep=0em,partopsep=0em]
    \item We propose a reference-free informative image captioning metric~\infometric. It can provide both coarse-grained scores and detailed token-level scores.
    \item We automatically construct training examples based on annotations in image caption datasets and design coarse- and fine-grained tasks to train the evaluation model.
    \item \infometric~achieves better correlation with human judgements on multiple benchmarks, as well as on our newly constructed fine-grained caption evaluation benchmark CapTokenEval.
\end{itemize}

\section{Related Work}

\noindent\textbf{Reference-only caption evaluation.} This type of evaluation only employs human-written captions as references and measures text similarity as the evaluation score.
Most widely used metrics such as \bleu~\cite{papineni2002bleu}, \rouge~\cite{lin2004rouge}, \meteor~\cite{banerjee2005meteor}, \cider~\cite{vedantam2015cider} and  \spice~\cite{anderson2016spice} all fall into this category.
\bleu~calculates the precision of n-gram matches; \rouge~measures the recall of the longest common subsequence; \meteor~utilizes wordnet-based synonym matching to relieve the shortage of exact word matching; \cider~introduces tf-idf to re-weight the importance of different n-grams; \spice~converts captions into scene graphs for similarity comparison.
One major limitation of the above metrics is that they cannot properly count synonym matches.
To overcome this deficiency, {\small BERT-S} \cite{DBLP:conf/iclr/ZhangKWWA20} leverages learned embeddings from a pretrained language model BERT \cite{kenton2019bert} to better measure semantic similarities.
{\small BERT-S++} \cite{DBLP:conf/acl/YiDH20} further improves {\small BERT-S} by taking into account the variance of multiple references.

\noindent\textbf{Reference+image caption evaluation.} 
As an image is worth a thousands of words, a limited number of references cannot fully cover image contents, making the reference-only caption evaluation less reliable.
Therefore, some works combine both references and images to evaluate generated captions. 
REO \cite{jiang2019reo} uses a pretrained image-text retrieval model SCAN \cite{lee2018stacked} to extract image contextualized caption features for computing relevance, extraness and omission scores.
TIGER \cite{jiang2019tiger} calculates grounding vectors for captions via SCAN to measure similarity, which represent how much captions are grounded in an image.
ViLBERTScore \cite{lee2020vilbertscore} is similar to BERT-S except that it generates visually-grounded features for each caption token by ViLBERT \cite{DBLP:conf/nips/LuBPL19}.
FAIEr \cite{DBLP:conf/cvpr/WangYWWC21} fuses scene graphs of the image and references as a union scene graph and compares it with the scene graph of generated captions. 

\begin{figure*}
    \centering
    \includegraphics[width=1.0\linewidth]{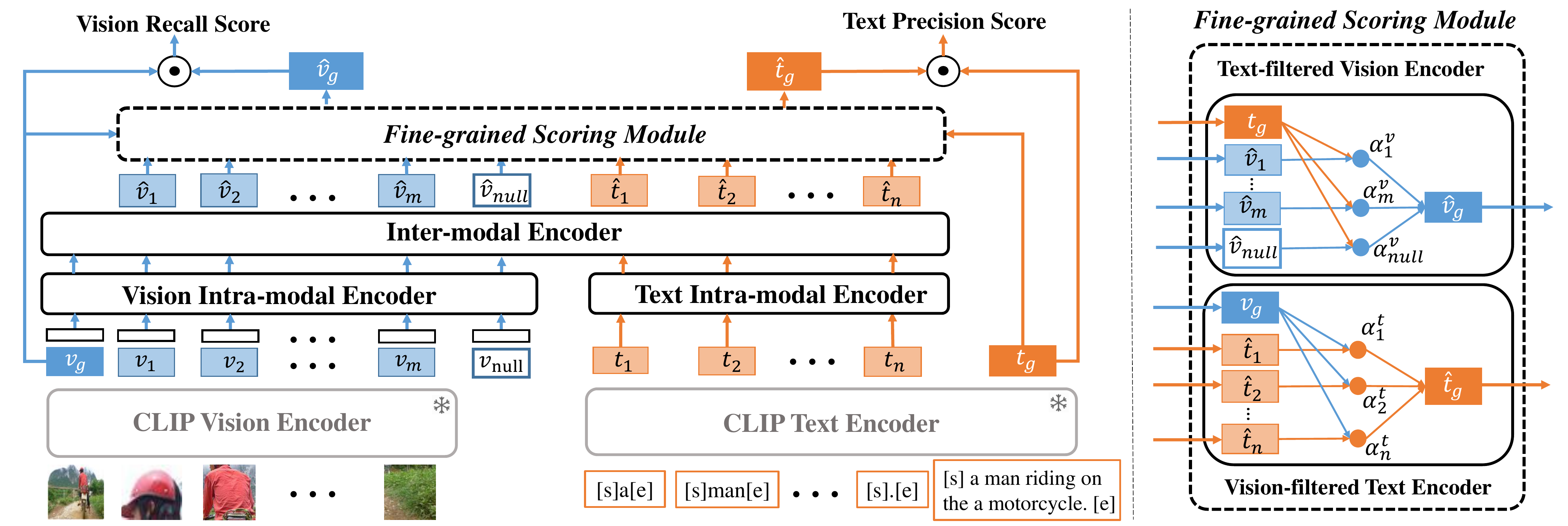}
    \caption{\textbf{Left:} the overall architecture of \textbf{Info}rmative \textbf{Met}ric for Reference-free \textbf{I}mage \textbf{C}aption evaluation (\infometric). \textbf{Right:} the detailed structure of the \emph{Fine-grained Scoring Module}.}
    \label{fig:overall_model}
\end{figure*}

\noindent\textbf{Reference-free caption evaluation.} 
To alleviate the annotation burden of obtaining references, a few works propose to evaluate image captions without references.
UMIC~\cite{DBLP:conf/acl/Lee0DBJ20} fine-tunes a pretrained multimodal transformer UNITER \cite{chen2020uniter} by contrastive learning to compute an image-text matching score. 
CLIP-S \cite{DBLP:conf/emnlp/HesselHFBC21} directly utilizes image-text similarity from CLIP \cite{DBLP:conf/icml/RadfordKHRGASAM21} - an image-text matching model trained on large-scale open-domain data. CLIP-S has achieved state-of-the-art evaluation performance.
However, these methods only provide single scores which are less informative to evaluate image captions.
In this work, we aim to provide more fine-grained feedbacks, not only indicating the captioning quality from precision and recall aspects, but also pointing out detailed mistakes such as incorrect words and unmentioned regions.

\section{Method}

We first introduce our model architecture in Sec~\ref{sec:method_model} and then describe the training and inference approaches in Sec~\ref{sec:method_train} and Sec~\ref{sec:method_inference} respectively.

\subsection{Model Architecture}
\label{sec:method_model}
Figure~\ref{fig:overall_model} illustrates the overall framework of our informative evaluation model, which consists of three modules: \emph{Token-level Encoding}, \emph{Intra\&Inter Modality Fusion} and \emph{Fine-grained Scoring}. 
Given an image $I$ and a caption $C$ as inputs, the Token-level Encoding module firstly generates a sequence of token-level features to represent the image and caption respectively.
Then the Intra\&Inter Modality Fusion module captures the intra- and inter-modality relationships.
Finally, the Fine-grained Scoring module produces token-level scores for each visual and textual token and derives vision recall, text precision, and overall scores based on the token-level scores.

\subsubsection{Token-level Encoding}
VLP models have shown superior performance and generalization ability in many vision-and-language tasks \cite{chen2020uniter}. Therefore, we utilize a state-of-the-art VLP model CLIP to extract token-level image and caption features. To be noted, our method can be adapted to different VLP models.

\noindent\textbf{Image Token Features.}
In order to obtain semantically meaningful image tokens, we use a pretrained object detector to detect region bounding boxes in image $I$. We encode each cropped region via CLIP vision encoder to get fine-grained token-level features $(v_{1}, ..., v_{m})$, where $m$ is the number of detected regions.
The whole image is encoded as a global vision feature $v_{g}$.
We further utilize a zero vector to represent a vision null token $v_{null}$, which aims to align with any texts irrelevant to the image.

\noindent\textbf{Caption Token Features.}
For a caption $C$, CLIP text encoder can generate a global feature $t_{g}$ to capture overall semantics of the whole sentence.
Although it could also generate a sequence of text token features, these features can overuse the sentence context, which harms fine-grained evaluation. An illustration about the context overuse can be found in Appendix \ref{context_overuse_issue}.
Therefore, we encode each token in $C$ separately as shown in Figure~\ref{fig:overall_model} to obtain independent token-level features $(t_{1},...,t_{n})$, where $n$ is the number of text tokens.

\vspace{-8pt}
\subsubsection{Intra\&Inter Modality Fusion}
In order to learn intra-modal relationships, we utilize two multi-layer transformers \cite{vaswani2017attention} to encode image and text tokens separately.
As spatial information is essential to infer relationships across image regions, we apply a linear layer to convert normalized bounding boxes as position features and add them to the initial image token features before fed into the intra-modal transformer.
Likewise, we add learnable position features for the text tokens.
For visual intra-modal encoding, we concatenate $v_{g}$ with $(v_1, \cdots, v_m, v_{null})$ to alleviate possible vision context loss in fine-grained image tokens due to imperfect detection. For textual intra-modal encoding, we directly utilize $(t_1, \cdots, t_m)$ tokens as inputs. 

We concatenate the image and text token-level features after intra-modal encoding and utilize an inter-modal encoder to learn correlation between vision and text modalities.
The inter-modal encoder is implemented as a multi-layer cross-modal transformer \cite{chen2020uniter}.
We denote the output features for image tokens as $\hat{V}=(\hat{v}_{1} ..., \hat{v}_{m}, \hat{v}_{null})$, output features for text tokens as $\hat{T}=(\hat{t}_{1}, ..., \hat{t}_{n})$.

\subsubsection{Fine-grained Scoring}
The Fine-grained Scoring module aims to predict which text tokens are incorrect and which image tokens are not mentioned. 
It consists of two cross-modal attention layers, namely Text-filterd Vision Encoder and Vision-filterd Text Encoder as shown in the right of Figure~\ref{fig:overall_model}. 
To identify which image tokens are mentioned, we use global text feature $t_g$ as query and token-level vision features $\hat{V}$ as key in the cross-modality attention layer to calculate visual token-level scores $\alpha^{v}$:
\begin{gather}
    s^{v}_{i} = (t_{g}W_{q}^{v})^{{\rm T}}\hat{v}_i W_{k}^{v},\label{eq:cross-attention}\\
    \alpha^{v} = {\rm Softmax}([s^{v}_{1}, ..., s^{v}_{m}, s^{v}_{null}]).
    \label{eq:token-level score}
\end{gather}
Similarly, to identify which text tokens are incorrect, we use global vision feature $v_g$ as query and token-level text features $\hat{T}$ as key to calculate textual token-level scores $\alpha^{t}$ by another cross-modality attention layer. 

Based on token-level scores, we derive vision recall score and text precision scores to measure the comprehensiveness and accuracy of generated captions respectively. 
We take visual token-level scores $\alpha^{v}$ and token-level vision features $\hat{V}$ to obtain a text-conditioned vision feature $\hat{v}_g$ by weighed average as follows:
\begin{gather}
    \hat{v}_g = \sum_{k \in \{1,...,m,null\}} \alpha^{v}_{k}\hat{v}_{k}.
\end{gather}
The more image regions are mentioned in a caption, the closer its text-conditioned vision feature should be to the global vision feature $v_g$. 
Thus, we compute the vision recall score as the cosine similarity between $\hat{v}_g$ and $v_g$, represented as $f^{R}(I, C) = \cos(\hat{v}_g, v_g)/\tau$, where $\tau$ is a learnable temperature parameter. Taking the untrained global vision feature $v_g$ as the comparison object, our vision recall score implicitly considers the salience of visual information, as illustrated in Appendix \ref{salience}.
In a similar way, we can obtain a vision-conditioned text feature $\hat{t}_g$ and compute a text precision score $f^{P}(I, C) = \cos(\hat{t}_g, t_g)/\tau$. 
Our overall score is the summation of precision score and recall score:
\begin{gather}
    f^{O}(I, C) = f^{R}(I, C) + f^{P}(I, C).
    \label{eqn:overall_score}
\end{gather}

\subsection{Multi-task Learning}
\label{sec:method_train}
To learn fine-grained token-level predictions as well as coarse-grained text precision and vision recall scores, we propose multiple training tasks to jointly optimize our evaluation model.

\subsubsection{Coarse-grained Score Learning}
Given an aligned image-caption pair $(I, C)$, we construct negative samples by pairing $I$ with other captions in the training batch or pairing $C$ with other images in the batch. Then, we calculate Noisy Contrastive Learning (NCE) loss $l_{r}$ based on vision recall scores and $l_{p}$ based on text precision scores. The NCE loss $l_{r}$ is calculated as follows:
\begin{gather}
l_{r} = (l^{i}_{r} + l^{c}_{r})/2, \label{eqn:loss_recall} \\
l^{i}_{r} = -\mathbb{E}_{(I,C)\sim B}\log \frac{e^{f^{R}(I, C)}}{\sum_{C' \in \mathcal{N}_{I}\cup\{C\}} e^{f^{R}(I, C')}}, \\
l^{c}_{r} = -\mathbb{E}_{(I,C)\sim B}\log \frac{e^{f^{R}(I, C)}}{\sum_{I' \in \mathcal{N}_{C}\cup\{I\}} e^{f^{R}(I', C)}},
\end{gather}
where $\mathcal{N}_{I}$ means a set of negative captions for image $I$ within the batch $B$, $\mathcal{N}_{C}$ means negative images for caption $C$.
The NCE loss $l_{p}$ is similar to Eq (\ref{eqn:loss_recall}) but utilizes $f^{P}(I, C)$ scores in computation.

\textbf{Hard Textual Negatives.} 
In the above coarse-grained score learning, negative captions for an image are randomly selected from the dataset and usually contains many irrelevant contents with the image. These textual negatives are not hard enough to learn a good vision recall score. Because the model could compute a high recall score for positive pairs by putting high weight to only one rather than all mentioned regions. 
To address this problem, we further design Hard Textual Negatives (HTN) during coarse-grained score learning. For multiple annotated captions of an image, we consider the one with more semantic words (nouns, verbs, adjectives and adverbs) should get higher vision recall score than the others. Therefore, we treat the other ones as hard textual negatives. The HTN loss $l^{h}_{r}$ is calculated as follows:
\iffalse
\begin{gather}
    l^{h}_{r} = -\mathbb{E}_{(i,c)\sim B}\log \frac{\exp(f^{R}(I, C))}{\mathcal{Z}_{I}^{h}}, \\
    \mathcal{Z}_{I}^{h} = \exp(f^{R}(I, C)) + \exp(f^{R}(I, C^{h})),
\end{gather}
\fi
\begin{gather}
    l^{h}_{r} = -\mathbb{E}_{(I, C)\sim B}\log \frac{e^{f^{R}(I, C)}}{e^{f^{R}(I, C)} + e^{f^{R}(I, C^{h})}},
\end{gather}
where $C^{h}$ is a hard textual negative for caption $C$.

\subsubsection{Fine-grained Score Learning}
To improve fine-grained evaluation, we design a sequence labeling task called Fine-grained Score learning.
We automatically generate supervision signals to learn token-level predictions.
For the text part, we prepare labels in a self-supervised manner. Given an image $I$ and its groundtruth caption $C$, we generate a polluted caption $C^{’}$ by randomly replacing a semantic word with a frequent word of the same part-of-speech tag. The text sequence label $Y^{t}$ for $(I, C^{’})$ is constructed by setting the polluted word as 0 (incorrect) and other semantic words as 1 (correct).
Non-semantic words such as adpositions, conjunctions are excluded in training. 
For the image part, we make use of existing phrase grounding annotations which align each phrase in a caption with its corresponding bounding boxes in the image.
The vision sequence label $Y^{v}$ for $(I, C)$ is constructed by setting all regions mentioned by the caption as 1 and otherwise 0. 

We use cross-entropy losses for both textual and visual fine-grained score learning tasks:
\begin{gather}
    l^{token}_{t} = -\frac{1}{n^{s}}\sum Y^{t}\log(\alpha^{t}), \\
    l^{token}_{v} = -\frac{1}{m}\sum Y^{v}\log(\alpha^{v}), 
\end{gather}
where $l^{token}_{t}$ and $l^{token}_{v}$ refer to the text-part and vision-part loss respectively, $\alpha^{t}$ and $\alpha^{v}$ are textual token-level scores and visual token-level scores in Eq (\ref{eq:token-level score}), $n^{s}$ is the number of semantic words.

\begin{table*}
    \caption{Overall score comparison on Flickr8k-Expert (F-Ex), Flickr8k-CF (F-CF), Composite (Com) and Pascal-50S. `w/ ref' means using 4-5 ground-truth references. `w/o ref'  means using no reference.}
    \label{tab:global score}
    \vspace{-10pt}
    %\footnotesize
    \small
    \centering
    \begin{tabular}{claaacccca}
    \toprule
    \multirow{2}*{\textbf{Type}} &
    \multirow{2}*{\textbf{Metric}} &  &  &  & \multicolumn{5}{c}{\textbf{Pascal-50S (accuracy)}} \\ 
    ~ & ~  & \multirow{-2}*{\textbf{F-Ex($\tau_{c}$)}} & \multirow{-2}*{\textbf{F-CF($\tau_{b}$)}} & \multirow{-2}*{\textbf{Com($\tau_{c}$)}} & HC & HI & HM & MM & Mean \\ 
    \midrule
    \multirow{11}*{w/ ref} & BLEU-4 & 30.8 & 16.9 & 30.6 & 52.5 & 90.4 & 63.0 & 42.3 & 55.8 \\ 
    ~ & ROUGE-L & 32.3 & 19.9 & 32.4 & 55.0 & 95.3 & 93.1 & 58.7 & 75.5 \\ 
    ~ & METEOR & 41.8 & 22.2 & 38.9 & 59.0 & 97.7 & 93.9 & 62.0 & 78.2 \\
    ~ & CIDEr & 43.9 & 24.6 & 37.7 & 53.7 & 98.1 & 90.8 & 63.7 & 76.6 \\
    ~ & SPICE & 44.9 & 24.4 & 40.3 & 56.9 & 96.3 & 87.1 & 66.4 & 76.7 \\
    ~ & BERT-S & 39.2 & 22.8 & 30.1 & 54.4 & 96.1 & 94.3 & 56.4 & 75.3 \\
    ~ & BERT-S++ & 46.7 & - & 44.9 & 65.4 & 98.1 & 96.4 & 60.3 & 80.1 \\
    ~ & TIGEr & 49.3 & - & 45.4 & 56.0 & 99.8 & 92.8 & 74.2 & 80.7 \\
    ~ & ViLBERTScore-F &  50.1 & - & 52.4 & 49.9 & 99.6 & 93.1 & 75.8 & 79.6\\
     ~ & FAIEr-4 & 52.6 & 35.4 & 57.7 & 59.7 & \textbf{99.9} & 92.7 & 73.4 & 81.4\\
    ~ & RefCLIP-S & 53.0 & 36.4 & 55.4 & 57.9 & 99.5 & 96.1 & 80.8 & 83.6 \\
    \midrule
    \multirow{6}*{w/o ref} & UMIC & 46.8 & - & 56.1 & 66.1 & 99.8 & 98.1 & 76.2 & 85.1 \\
   ~ & FAIEr-r & 50.1 & 32.4 & 50.5 & - & - & - & - & -\\ 
   ~ & CLIP-S & 51.5 & 34.4 & 53.8 & 60.4 & 99.4 & \textbf{97.8} & 77.1 & 83.7 \\
   ~ & CLIP-S$_{tune}$ & 54.3 & 36.6 & 57.3 & 61.0 & 99.5 & 95.9 & \textbf{82.0} & 84.6 \\
   ~ & InfoCLIP & 32.6 & 23.5 & 15.3 & 37.3 & 87.3 & 58.9 & 72.9 & 64.1 \\
   ~ & InfoCLIP$_{tune}$ & 37.7 & 27.7 & 24.6 & 37.3 & 92.5 & 62.7 & 74.7 & 66.8 \\
   ~ & \infometric & 54.2 & 36.3 & 59.2 & 69.0 & 99.8 & 94.0 & 78.3 & 85.3 \\
   ~ & \infometric$^{+}$ & \textbf{55.5} & \textbf{36.6} & \textbf{59.3} & \textbf{69.9} & 99.7 & 96.8 & 79.6 & \textbf{86.5} \\
    \bottomrule
    \end{tabular}
\end{table*}

\subsection{Inference}
\label{sec:method_inference}
Given input pair $(I, C)$, we first compute token-level scores $\alpha^{v}$ and $\alpha^{t}$ for fine-grained prediction with a threshold $\beta$. Considering that a caption hardly contains more than 10 semantic words, we set $\beta$ as 0.1. For the text part, semantic tokens with a score greater than $\beta$ are judged as correct ones. For the image part, regions with a score greater than $\beta$ are identified as mentioned ones.

Then we calculate the vision recall, text precision, and overall scores as in Eq (\ref{eqn:overall_score}). We denote our vision recall score $f^{R}(I,C)$ as \infometric$^{R}$, text precision score $f^{P}(I,C)$ as \infometric$^{P}$, and overall score $f^{O}(I,C)$ as \infometric.
Furthermore, we combine our overall score with the CLIP similarity:
\begin{gather}
     {\rm \infometric}^{+} = {\rm \infometric} + \frac{\cos(v_g, t_g)}{\tau^{clip}}
\end{gather}
where $\tau^{clip}$ is the temperature of CLIP.

\section{Experiment}

\subsection{Experimental Setting}
\noindent\textbf{Training Datasets.}
With the training splits of Flickr30k \cite{young2014image} and MSCOCO \cite{lin2014microsoft} datasets,
we construct 715,662 image-caption pairs for general coarse-grained score learning, and 611,105 triplets with hard textual negatives. 
For fine-grained score leaning, we construct 512,000 samples from MSOCO and Flick30k for the text part training and 178,689 samples from Flickr30k for the vision part training.

\noindent\textbf{Implementation Details.}
We use CLIP(ViT-B/32) for token-level encoding. The image regions are detected by the bottom-up model \cite{DBLP:conf/cvpr/00010BT0GZ18}. To remove redundant bounding boxes, we use k-means algorithm to generate 20 clusters among 100 detected regions and select one region per cluster. The details can be found in Appendix \ref{cluster_details}. The maximum length for textual tokens is set as 32.
In the intra\&inter modality fusion, intra- and inter-modal encoders contain 4 and 2 transformer layers respectively.
During training, the batch size is set as 32 and the initial learning rate is set as 1e-4. We iteratively train our model on multiple tasks for 32,000 iterations. The training ratio of coarse- and fine-grained tasks is 3:1. The training takes 5 hours on 4 V100 GPUs.

\subsection{Coarse-grained Score Evaluation}
\subsubsection{Evaluation Datasets} 
\noindent\textbf{Flickr8k-Expert}~\cite{DBLP:journals/jair/HodoshYH13} contains 5,644 pairs of images and machine-generated captions. 
Each pair is scored from 1 (irrelevant) to 4 (well related) by 3 expert annotators.

\noindent\textbf{Flickr8k-CF}~\cite{DBLP:journals/jair/HodoshYH13} consists of 47,830 image-captions pairs. Each pair is judged "yes" or "no" by at least 3 annotators, where "yes" is for good captions. The final score of each pair is determined by the proportion of "yes". 

\noindent\textbf{Composite} \cite{DBLP:journals/cviu/AdityaYBAF18} contains 3,995 images from MSCOCO, Flickr30K and Flickr8k \cite{hodosh2013framing}. For each image, there are two machine-generated captions and one human-written caption. Every image-caption pair is scored from 1 (irrelevant) to 5 (perfectly related). 

\noindent\textbf{Pascal-50S} \cite{DBLP:conf/cvpr/VedantamZP15} contains 4,000 triplets, each of which contains an image and two captions. Annotators are asked to judge which caption is better. According to caption types, Pascal-50S is evenly split into 4 subsets: `HC' means two correct human-written captions; `HI' means two human-written captions but one is wrong; `HM' means one human-written caption and one machine-generated caption; `MM' means two machine-generated captions.

\noindent\textbf{THumB 1.0} \cite{kasai2021transparent} contains 500 images from MSCOCO. Each image is paired with one human-written caption and four machine-generated captions. For each image-caption pair, there are a precision score measuring the accuracy of the caption, a recall score assessing how much of the salient information is covered, and a total score measuring the overall quality. 

\subsubsection{Evaluation Metrics} 
We follow previous works \cite{DBLP:conf/emnlp/HesselHFBC21,DBLP:conf/cvpr/VedantamZP15,kasai2021transparent} to evaluate captioning metrics.
We use kendall-c correlation ($\tau_{c}$) on Flickr8k-Expert, kendall-b correlation ($\tau_{b}$) on Flickr8k-CF, kendall-c correlation ($\tau_{c}$) on Composite, classification accuracy on Pascal-50s and Pearson correlation ($\rho$) on THumB 1.0.

\begin{table*}
    \caption{Ablation Study on Flickr8k-Expert (F-Ex), Flickr8k-CF (F-CF), Composite (Com), Pascal-50S and THumB 1.0. HTN denotes using hard text negatives in coarse-grained score learning, and FS refers to the fine-grained score leaning. `$v_{g}$' means incorporating the vision global feature in the Intra\&Inter Modality Fusion Module.}
    \label{tab:ablation}
    \vspace{-10pt}
    \tabcolsep=0.12cm
    \small
    \centering
    \begin{tabular}{ccccccaaaccccaccacca}
    \toprule
      \multirow{2}*{\textbf{Id}} & \multicolumn{3}{c}{\textbf{Architecture}} & \multicolumn{2}{c}{\textbf{Training}} &  &  &  & \multicolumn{5}{c}{\textbf{Pascal-50S}} & \multicolumn{3}{c}{\textbf{THumB w/o h}} & \multicolumn{3}{c}{\textbf{THumB w/ h}} \\ 
    ~ & Intra & Inter & $v_{g}$ & HTN & FS & \multirow{-2}*{\textbf{F-Ex}}  & \multirow{-2}*{\textbf{F-CF}} & \multirow{-2}*{\textbf{Com}} & HC & HI & HM & MM & Mean & P & R & Total & P & R & Total \\ 
    \midrule
    r1 & $\checkmark$ & & $\checkmark$ &  &  &  51.7 & 36.8 & 57.8 & 58.0 & 99.5 & 95.0 & 76.3 & 82.2 & 0.23 & 0.26 & 0.35 & 0.20 & 0.26 & 0.32 \\
    r2 &  & $\checkmark$ & $\checkmark$ &  &  &  55.1 & \textbf{37.1} & 59.0 & 59.5 & 99.8 & 95.4 & 78.1 & 83.2 & 0.23 & 0.26 & 0.35 &  0.20 & 0.26 & 0.32 \\
    r3 & $\checkmark$ & $\checkmark$ & &  &  &   55.1 & 36.9 & \textbf{59.4} & 58.6 & \textbf{99.9} & 95.7 & 79.6 & 83.5 & 0.21 & 0.26 & 0.34 &  0.19 & 0.26 & 0.32 \\
    r4 & $\checkmark$ & $\checkmark$ & $\checkmark$ &  &  &   \textbf{55.2} & 36.9 & 59.3 & 58.0 & 99.7 & 96.1 & \textbf{80.8} & 83.7 & 0.22 & 0.26 & 0.35 &  0.20 & 0.26 & 0.33 \\
    r5 & $\checkmark$ & $\checkmark$ & $\checkmark$ & $\checkmark$ &  &  54.5 & 36.2 & 58.8 & \textbf{69.3} & 99.6 & 93.7 & 75.2 & 84.5 & \textbf{0.23} & 0.28 & \textbf{0.37} &  \textbf{0.22} & 0.30 & 0.37 \\
    r6 & $\checkmark$ & $\checkmark$ & $\checkmark$  & & $\checkmark$ &\textbf{55.2} & 37.0 & 59.3 & 60.2 & 99.7 & 96.8 & 79.6 & 84.1 & 0.22 & 0.26 & 0.34 &  0.20 & 0.26 & 0.32 \\
    r7 & $\checkmark$ & $\checkmark$ & $\checkmark$ & $\checkmark$ & $\checkmark$ &  54.2 & 36.3 & 59.2 & 69.0 & 99.8 & 94.0 & 78.3 & \textbf{85.3} & 0.22 & \textbf{0.30} & \textbf{0.37} &  0.21 & \textbf{0.32} & \textbf{0.38} \\
    \bottomrule
    \end{tabular}
\end{table*}

\begin{table}
    \caption{Experiments on THumB 1.0. `w/o Human' means discarding human annotated image-caption pairs.}
    \label{tab:THumB}
    \vspace{-10pt}
    %\footnotesize
    \small
    \centering
    \begin{tabular}{p{0.02\linewidth}p{0.20\linewidth}p{0.03\linewidth}p{0.07\linewidth}p{0.07\linewidth}|p{0.03\linewidth}p{0.07\linewidth}p{0.07\linewidth}}
    \toprule
    \multirow{2}*{\textbf{Ref}} & \multirow{2}*{\textbf{Metric}}  &  \multicolumn{3}{c|}{\textbf{w/o Human}} &
    \multicolumn{3}{c}{\textbf{w/ Human}}\\ 
    ~ & ~ & P & R & Total & P & R & Total \\
    \midrule
    \multirow{6}*{w/} & BLEU & .21 & .13 & .25 & .15 & .04 & .13 \\
    ~ & ROUGE-L & .26 & .17 & .31 & .18 & .07 & .18 \\
    ~ & CIDEr & .27 & .18 & .33 & .21 & .11 & .23 \\
    ~ & SPICE & .26 & .15 & .30 & .20 & .09 & .21 \\
    ~ & BERT-S & .27 & .18 & .33 & .20 & .10 & .21 \\
    ~ & RefCLIP-S & \textbf{.34} & \textbf{.27} & \textbf{.44} & \textbf{.31} & \textbf{.26} & \textbf{.41} \\
    \midrule
    \multirow{12}*{w/o} & InfoCLIP$^{R}$ & .05 & .19 & .17 & .05 & .19 & .17 \\
    ~ & InfoCLIP$^{P}$ & .11 & -.22 & -.08 & .09 & -.20 & -.08 \\
    ~ & InfoCLIP & .13 & -.06 & .04 & .11 & .06 & .03 \\
    ~ & InfoCLIP$_{tune}$ & .15 & -.15 & .00 & .11 & -.15 & -.03 \\
    ~ & CLIP-S & .18 & .27 & .32 & .17 & .28 & .32 \\
    ~ & CLIP-S$_{tune}$ & .15 & .26 & .29 & .13 & .26 & .28 \\
    ~ & \infometric$^{R}$ & .18 & .29 & .34 & .19 & .32 & .36 \\  
    ~ & \infometric$^{P}$ & .23 & .27 & .36 & .20 & .27 & .33 \\
    ~ & \infometric & .22 & .30 & .37 & .21 & .32 & .38 \\
    ~ & \infometric$^{+}$ & \textbf{.22} & \textbf{.33} & \textbf{.39} & \textbf{.21} & \textbf{.34} & \textbf{.39} \\
    \bottomrule
    \end{tabular}
\end{table}

\vspace{-8pt}
\subsubsection{Comparison with State of the Arts}
We compare \infometric~with SOTA methods as well as three strong baselines: CLIP-S$_{tune}$, InfoCLIP and InfoCLIP$_{tune}$. 
CLIP-S$_{tune}$ calculates an overall score as CLIP-S \cite{DBLP:conf/emnlp/HesselHFBC21} but is fine-tuned on MSCOCO and Flickr30k. InfoCLIP directly uses CLIP to perform fine-grained scoring like \infometric~but removes the Intra\&Inter Modality Fusion and parameters in Fine-grained Scoring. InfoCLIP$_{tune}$ is a fine-tuned version of InfoCLIP.
More details can be found in the Appendix \ref{baseline_details}.

Table \ref{tab:global score} shows the overall score comparison on Flickr8k-Expert, Flickr8k-CF, Composite and Pascal-50S.
Our reference-free metric \infometric~achieves state-of-the-art correlation with human judgements on Composite and Pascal-5OS. It is on par with the strong baseline CLIP-S$_{tune}$ on Flickr8k-Expert and Flickr8k-CF. To be noted, \infometric~performs much better than InfoCLIP, which proves the necessity of our model architecture upon CLIP backbones.  After combined with CLIP similarity, \infometric$^{+}$ further improves performances on all benchmarks. 

To separately evaluate the performance of our vision recall score \infometric$^{R}$ and text precision score \infometric$^{P}$, we further conduct experiments on THumB 1.0 in Table \ref{tab:THumB}.
\textbf{First}, by comparing \infometric$^{P}$ and \infometric$^{R}$, \infometric$^{R}$ achieves better correlation with human-labeled recall score and \infometric$^{P}$ achieves better correlation with human-labeled precision score. This indicates that our \infometric$^{R}$ and \infometric$^{P}$ indeed evaluates the recall of image contents and the precision of caption respectively. Besides, both \infometric$^{P}$ and \infometric$^{
R}$ surpass the state-of-the-art reference-free metric CLIP-S on total score correlation.
\textbf{Second}, our overall score \infometric~achieves significant boost on total score, which demonstrates that precision and recall are complementary in human's final evaluation for captions. \infometric$^{+}$ slightly improves the total score performance.
\textbf{Third}, compared with the state-of-the-art reference-based metric RefCLIP-S \cite{DBLP:conf/emnlp/HesselHFBC21}, our \infometric$^{+}$ achieves much better recall correlation but lower precision correlation with humans. This is because text-text semantic comparison is much easier than cross-modal semantic comparison, making the precision correlation of reference-based metrics higher. However, limited textual references cannot fully capture image contents, which is harmful for vision recall.
\textbf{Finally}, \infometric~ achieves much better performance than InfoCLIP, which shows the effectiveness of our proposed modules on top of CLIP.

\subsubsection{Ablation Study}
We first validate the effectiveness of our model architecture.
As shown in Table \ref{tab:ablation}, removing Intra-modal encoders (r2 vs r4) or Inter-modal encoder (r1 vs r4) results in performance drop on Flickr8k-Expert, Composite and Pascal-50S. Besides, removing global vision feature $v_{g}$ from Intra\&Inter encoding (r3 vs r4) leads to slight performance drop on Flickr8k-Expert, Pascal-50S and THumB1.0.

We then carry out ablation study to verify the effectiveness of our training strategy in Table \ref{tab:ablation}. 
Our proposed hard textual negatives (r4 vs r5) achieves significant improvements on HC subset of Pascal50s and THumB 1.0 Recall. This shows that constructing hard negatives indeed helps model better evaluate the vision content recall.
Adding fine-grained score learning task (r4 vs r6) is also beneficial to the performance of coarse-grained score, which performs better on Pascal-50S and is comparable on other datasets.
When trained with all tasks together (r7), \infometric~further improves on Pascal-50S and THumB 1.0, and achieves state-of-the-art performance on all datasets.

\begin{figure*}
    \centering
    \includegraphics[width=0.95\linewidth]{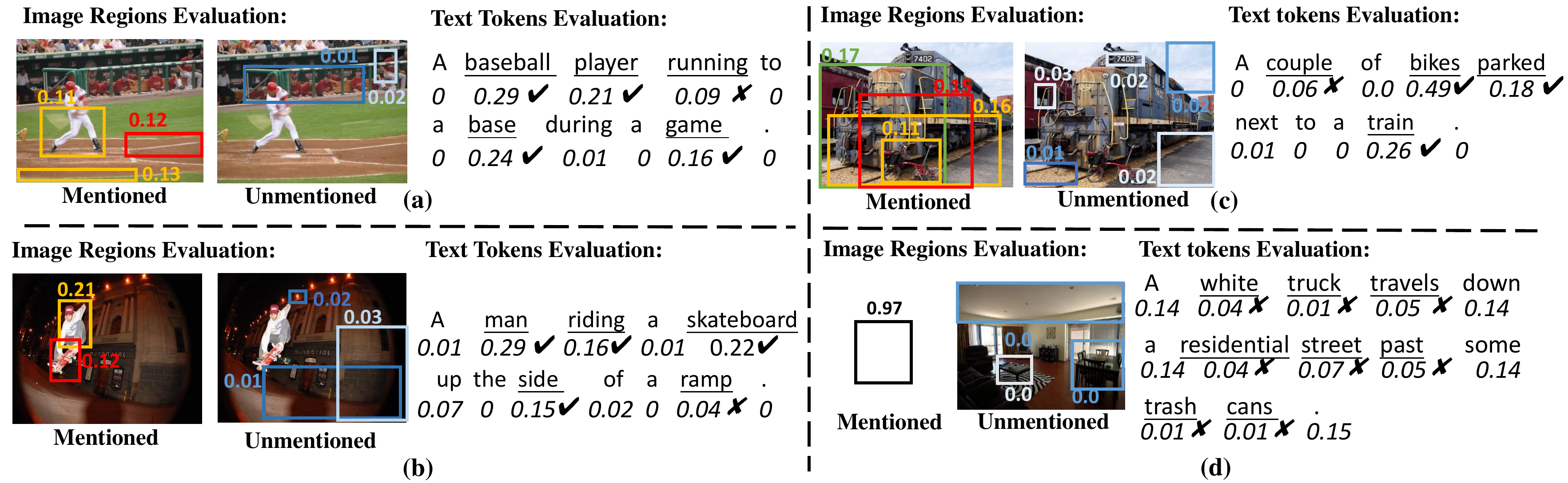}
    \vspace{-10pt}
    \caption{Fine-grained evaluation examples using \infometric. Semantic words in captions are underlined. Semantic words and image regions with a score lower than 0.1 are considered as incorrect words and unmentioned regions respectively. We show all mentioned regions but only the top unmentioned image regions for better visualization.}
    \label{fig:cases}
\end{figure*}

\subsection{Generalization Ability}
\begin{table}
    \caption{Cross-modal retrieval performances on Nocaps.}
    \label{tab:nocaps}
    \vspace{-10pt}
    %\footnotesize
    \small
    \centering
    \begin{tabular}{p{0.22\linewidth}p{0.05\linewidth}p{0.05\linewidth}p{0.1\linewidth}|p{0.05\linewidth}p{0.05\linewidth}p{0.1\linewidth}}
    \toprule
    \multirow{2}*{\textbf{Method}}  &   \multicolumn{3}{c|}{\textbf{image to text}} &
    \multicolumn{3}{c}{\textbf{text to image}}\\ 
    ~ & R@1 & R@5 & R@10 & R@1 & R@5 & R@10 \\
    \midrule
    TIGER & 63.8 & 87.0 & 92.4 & 22.5 & 66.5 & 81.9 \\
    CLIP-S & 88.2 & 98.3 & 99.7 & 67.5 & 91.5 & 95.8 \\
   \infometric  & 76.6 & 96.5 & 99.1 & 71.6 & 94.4 & 97.7 \\
   \infometric$^{+}$ & \textbf{90.9} & \textbf{98.8} & \textbf{99.7} & \textbf{76.2} & \textbf{95.9} & \textbf{98.4} \\
    \bottomrule
    \end{tabular}
\end{table}
\infometric~are trained with image-captions of Flick30k and MSCOCO. To evaluate its generalization ability, we further conduct experiments on NoCaps \cite{agrawal2019nocaps}, whose objects are greatly different from Flick30k and MSCOCO. 
Since there are no human-labeled scores for image-caption pairs, we perform text-image cross-modal retrieval to validate the effectiveness of our metric. 
As shown in Table \ref{tab:nocaps}, \infometric~performs worse than CLIP-S on image-to-text retrieval but better on text-to-image retrieval. After combining with CLIP similarity, \infometric$^{+}$ achieves the state-of-the-art performance on both two retrieval tasks. 
It indicates our overall score can also perform well on instances with unseen objects.

\subsection{Fine-grained Score Evaluation}
\noindent\textbf{Dataset.} 
To validate the token-level evaluation performance of \infometric, we collect a fine-grained caption evaluation benchmark called CapTokenEval\footnote{We will release codes and datasets upon acceptance.}. CapTokenEval is built upon a subset of THumB 1.0. We select 700 image-caption pairs whose precision scores are not perfect (< 5.0). For the text part, annotators are asked to judge which words are irrelevant with the image. For the image part, we collect 20 bounding boxes and ask annotators to identify mentioned regions. More details about the annotation can be found in Appendix~\ref{anno_details}.

\noindent\textbf{Quantitative Results.}
Given each image-caption pair, \infometric~produces sequence of prediction for both image regions and caption tokens. To quantify token-level evaluation performance, for the text part, we only calculate the accuracy of semantic tokens (nouns, verbs, adjectives and numbers).
As shown in Table \ref{tab:token score}, without extra parameters, InfoCLIP achieves promising performance for fine-grained visual evaluation but poor performance in the text part. 
Consistent with the result shown in Table \ref{tab:THumB} that InfoCLIP$^R$ ourperforms InfoCLIP$^P$, it further shows the importance of context fusion for text precision evaluation.
With multi-task learning, \infometric~achieves promising prediction accuracy on both vision and text sequence. Both hard textual negatives and fine-grained score learning task contribute to token-level evaluation performance. Notably, fine-grained score learning task greatly boosts the text-part accuracy. 
Coarse-grained contrastive learning for text precision score within a batch can result in the model only putting relatively higher weights on a few correct text tokens. Our fine-grained score learning task could effectively alleviate this lazy behavior by teaching the model to put high weights on all correct tokens.

\noindent\textbf{Qualitative Results.}
We show some qualitative results of token-level evaluation in Figure \ref{fig:cases}. Firstly, \infometric~is able to identify various mistakes made in captions, including wrong actions (e.g.``running'' in case a), wrong objects (e.g.``ramp'' in case b), and wrong modifiers (e.g.``couple'' in case c). Secondly, \infometric~could report mentioned image regions (e.g. the ``skateboard'' region in case b)  and unmentioned regions (e.g. the ``building'' region in case b). Especially, when the caption is totally irrelevant with the image, as shown in case d, \infometric~could not only judge the wrong semantic words but also inform that all image regions are not mentioned by putting a very high score to the vision null token.
One limitation of current metric is that although we perform region filtering by clustering, we still find some similar regions as shown in Figure \ref{fig:cases}(c). Better ways to de-duplicate image regions could bring further improvement.

\begin{table}
    \caption{Token-level evaluation on CapTokenEval. CS: coarse-grained score learning; HTN: adding hard textual negatives in CS; FS: fine-grained score leaning.}
    \label{tab:token score}
    \vspace{-10pt}
    %\footnotesize
    \small
    \centering
    \begin{tabular}{p{0.2\linewidth}p{0.07\linewidth}p{0.07\linewidth}p{0.07\linewidth}p{0.07\linewidth}p{0.07\linewidth}}
    \toprule
    \multirow{2}*{Method}  &  \multicolumn{3}{c}{Training} & \multicolumn{2}{c}{Accuracy}  \\
     ~ & CS & HTN & FS & Vision & Text \\
    \midrule
    InfoCLIP & - & - & - & 0.73 & 0.33  \\
    InfoCLIP$_{tune}$ & - & - & - & 0.74 & 0.37  \\
    \midrule
    \multirow{4}*{Ours}& \checkmark & $\times$ & $\times$ & 0.74 & 0.36 \\
    ~ & \checkmark &\checkmark & $\times$ & 0.75 & 0.37 \\
     ~ & \checkmark & $\times$ &\checkmark & 0.75 & 0.79 \\
    ~ &\checkmark & \checkmark & \checkmark & \textbf{0.75} & \textbf{0.80} \\
    \bottomrule
    \end{tabular}
\end{table}
\section{Conclusion}
To provide feedbacks on detailed mistakes of image captions, we propose a reference-free informative metric \infometric~based on a state-of-the-art vision-language model. \infometric~not only points out incorrect descriptions, but also tells which regions are not mentioned. Based on these fine-grained evaluation, \infometric~derives a text precision score, a vision recall score, and an overall score. We design both coarse- and fine-grained training tasks to optimize our metric. The overall score given by our metric achieves state-of-the-art correlation with human judgement on multiple benchmarks. We further build a token-level caption evaluation benchmark CapTokenEval to prove the effectiveness of our fine-grained evaluation. 
\section*{Limitations}
\label{limits}
This work focuses on informative image captioning evaluation, including an overall score, vision recall, text precision and token-level scores. The effectiveness of our metric is validated on standard image captioning benchmarks. \infometric~in this work may not perform well in other captioning tasks due to domain gap, but we contend that our general framework can be adapted to other domains such as text-aware image captioning. For example, for text-aware image captioning which focuses more on scene texts in images, we could further encode text regions besides the existing object regions for better comparison with captions. In the future, we will comprehensively explore how to adapt our metric to other captioning tasks, such as text-aware image captioning and video captioning.

\bibliography{anthology,reference}
\bibliographystyle{acl_natbib}

\appendix
\begin{figure*}[!tbp]
    \centering
    \includegraphics[width=1.0\linewidth]{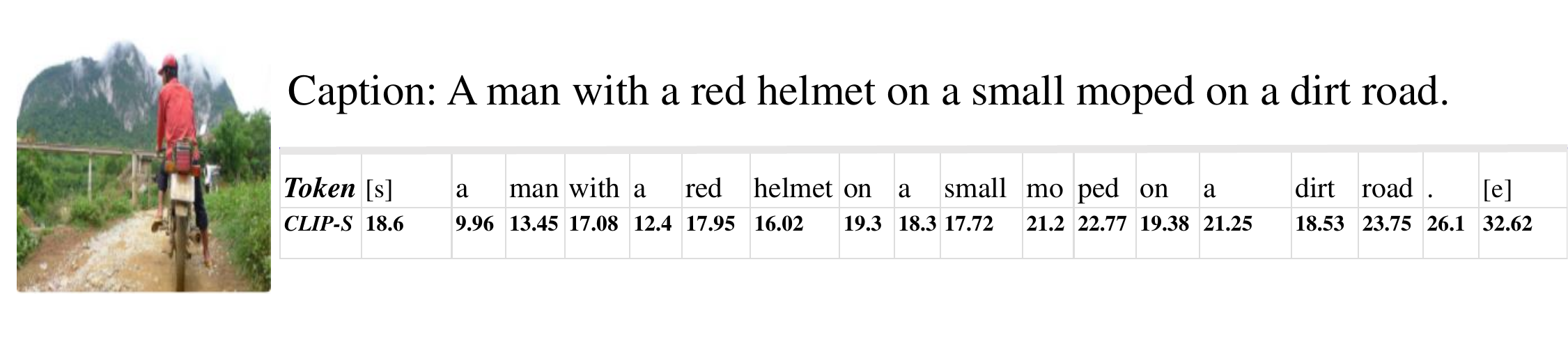}
    \caption{An illustration about the context overuse during text encoding of CLIP. The CLIP-S of each token are calculated with global vision feature and token-level text feature got by original CLIP encoding way rather than individually encoding. }
    \label{fig:context overuse}
\end{figure*}

\section{Context Overuse Issue}
\label{context_overuse_issue}
CLIP \cite{DBLP:conf/icml/RadfordKHRGASAM21} is trained to well align global image representations and sentence representation. Thus it applies a triangle masking during text encoding and treats the representation of the last text token [e] as the sentence representation. Due to the training objective and text masking mechanism, the text context information is accumulated with the sequence order, which is unfavorable for text-part fine-grained evaluation. As shown in Figure \ref{fig:context overuse}, the third `a' is a meaningless indefinite article but gets a higher relevance score than the correct noun `man'.

\begin{figure}[h]
    \centering
    \includegraphics[width=1.0\linewidth]{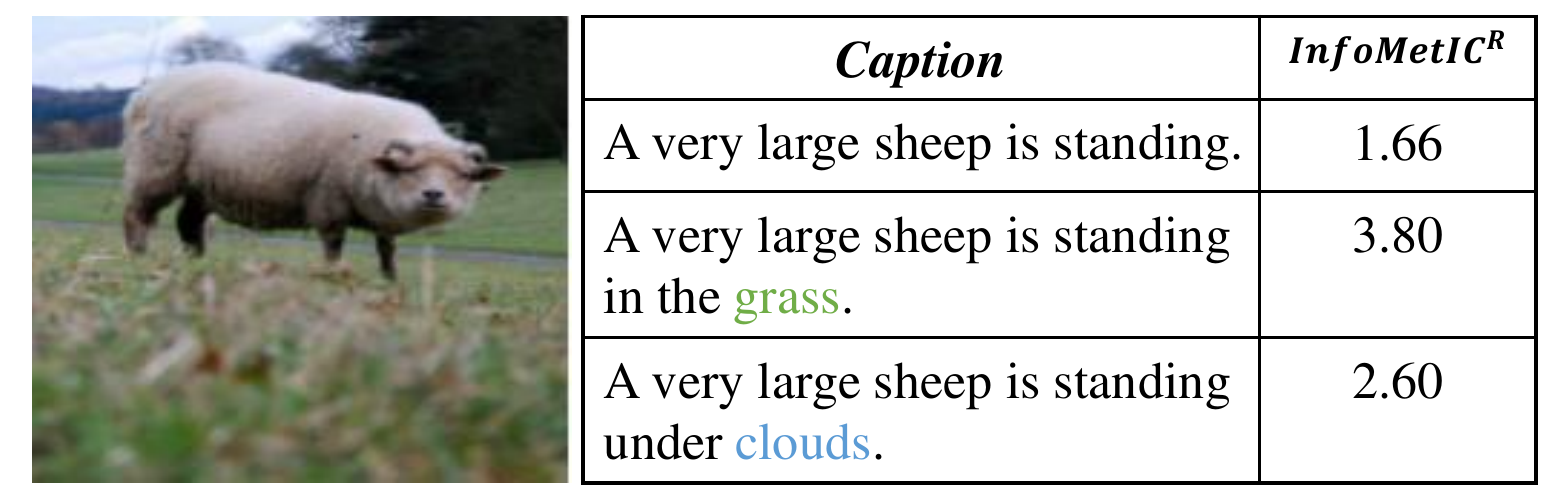}
    \caption{An illustration about the influence of object salience to our vision recall score \infometric$^{R}$.}
    \label{fig:salience}
\end{figure}

\section{Salience of Visual Information}
\label{salience}
Our vision recall score is calculated by comparing the text-conditioned vision features (the CLIP’s global vision feature) rather than the sum or average of all regions features. CLIP is trained with massive image-caption pairs and achieves promising performance on multiple Vision-Language tasks. Thus it’s convincing that the global vision feature produced by CLIP could well represent the salient information in an image. As illustrated in  Figure \ref{fig:salience}, both `cloud' and `grass' are objects in the image, but \infometric~gives the second caption higher vision recall score because `grass' is more salient than `clouds' in the image.

\section{Cluster Number Setting Details}
\label{cluster_details}
Similar image regions can cause confusion during fine-grained evaluation. In this work, redundant regions are removed by K-means clustering algorithm. Concretely, with 100 bounding boxes given by the object detection model, we perform K-means to generate $N$ clusters. For each cluster, the region with highest confidence score given by the object detection model is maintained. The evaluation performance of \infometric~with different $N$ settings is shown in Table \ref{tab:cluster setting}. With the cluster number ranging from 10 to 50, the overall evaluation performance of \infometric~shows minor difference on these benchmarks. Taking into account both performance and complexity, we finally set $N$ as 20. 

\begin{table}
    \caption{Performance of \infometric~with different cluster numbers on Flickr8k-Expert (F-Ex), Flickr8k-CF (F-CF), Composite (Com), Pascal-50S and THumB w/ Human.}
    \label{tab:cluster setting}
    \vspace{-10pt}
    %\footnotesize
    \small
    \centering
    \begin{tabular}{cccccc}
    \toprule
    cluster & F-Ex & F-CF & Com & Pascal50S & Thumb \\ 
    \midrule
     10 & 54.2 & 36.1 & 58.3 & 84.8 & 0.36 \\ 
     20 & 54.2 & 36.3 & 59.2 & 85.3 & 0.38 \\ 
     30 & 54.4 & 36.3 & 59.5 & 85.2 & 0.36 \\ 
     40 & 54.7 & 36.2 & 59.2 & 85.3 & 0.39 \\ 
     50 & 54.8 & 36.3 & 59.5 & 85.3 & 0.37 \\ 
    \bottomrule
    \end{tabular}
\end{table}

\begin{figure*}
    \centering
    \includegraphics[width=1.0\linewidth]{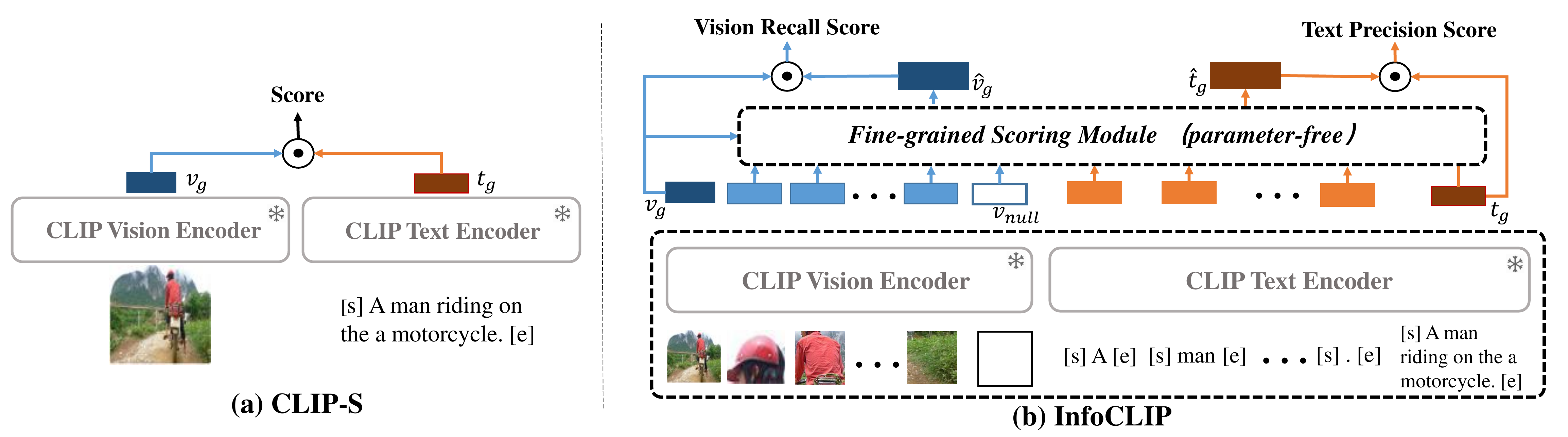}
    \caption{Overall architectures of baselines CLIP-S and InfoCLIP}
    \label{fig:baselines}
\end{figure*}

\section{Baseline Details}
\label{baseline_details}
To verify the effectiveness of \infometric, besides state-of-the-art caption metrics, we set extra three baselines CLIP-S$_{tune}$, InfoCLIP and InfoCLIP$_{tune}$. As shown in Figure \ref{fig:baselines}(a), CLIP-S \cite{DBLP:conf/emnlp/HesselHFBC21} directly uses the global representations given by CLIP\cite{DBLP:conf/icml/RadfordKHRGASAM21} to calculate a cosine similarity as the overall score.
CLIP-S$_{tune}$ follows the same calculation manner but uses a CLIP fine-tuned on MSCOCO and Flickr30k as the backbone. Previous metrics can't do fine-grained caption evaluation. Therefore, we set a fine-grained evaluation baseline InfoCLIP, as shown in Figure \ref{fig:baselines}(b). InfoCLIP performs fine-grained scoring as \infometric~without Intra\&Inter Modality Fusion and parameters in Fine-grained Scoring, e.g.$W^{v}_{q}$ and $W^{v}_{k}$ in Eq (\ref{eq:cross-attention}).  InfoCLIP$_{tune}$ means using a fine-tuned CLIP as the backbone.

\section{CapTokenEval Annotation Details}
\label{anno_details}
To quantify caption evaluation performance at token level, we collect a fine-grained caption evaluation benchmark called CapTokenEval. The details of our annotation are introduced in following subsections.

\subsection{Data Preparation}
\label{prepration}
We prepare image-caption pairs for annotation based on the publicly released dataset THumB 1.0 \cite{kasai2021transparent}.  
THumB 1.0 collects 500 images from MSCOCO \cite{lin2014microsoft} and pairs each image with 4 captions generated by state-of-the-art image captioning models, including UP-Down \cite{DBLP:conf/cvpr/00010BT0GZ18}, Unified-VLP \cite{DBLP:conf/aaai/ZhouPZHCG20}, VinVL-base and VinVL-large \cite{zhang2021vinvl}. There are a precision score, a recall score and a total score for each image-caption pair. To ensure that textual token-level evaluation in our benchmark is hard enough, we select image-caption pairs whose precision score is not perfect (<5.0). We finally collect 700 image-captions pairs from ThumB 1.0. As the data used in our annotation all come from publicly released datasets, there are no ethic issues.

For each image, we extract 100 bounding boxes with pre-trained object detection model Bottom-Up \cite{DBLP:conf/cvpr/00010BT0GZ18}. To filter similar image regions, we apply K-means clustering on these bounding boxes. We generate 20 clusters for each image and choose a bounding box with highest confidence score of object classification from each cluster. Thus, for each image-caption pair, we provide 20 image regions to annotators, who will choose which regions are mentioned by the caption. For the text part, we tokenize the caption with Spacy\footnote{https://spacy.io/usage}.

\subsection{Annotation Platform}
\label{tool}
\begin{figure*}
    \centering
    \includegraphics[width=1.0\linewidth]{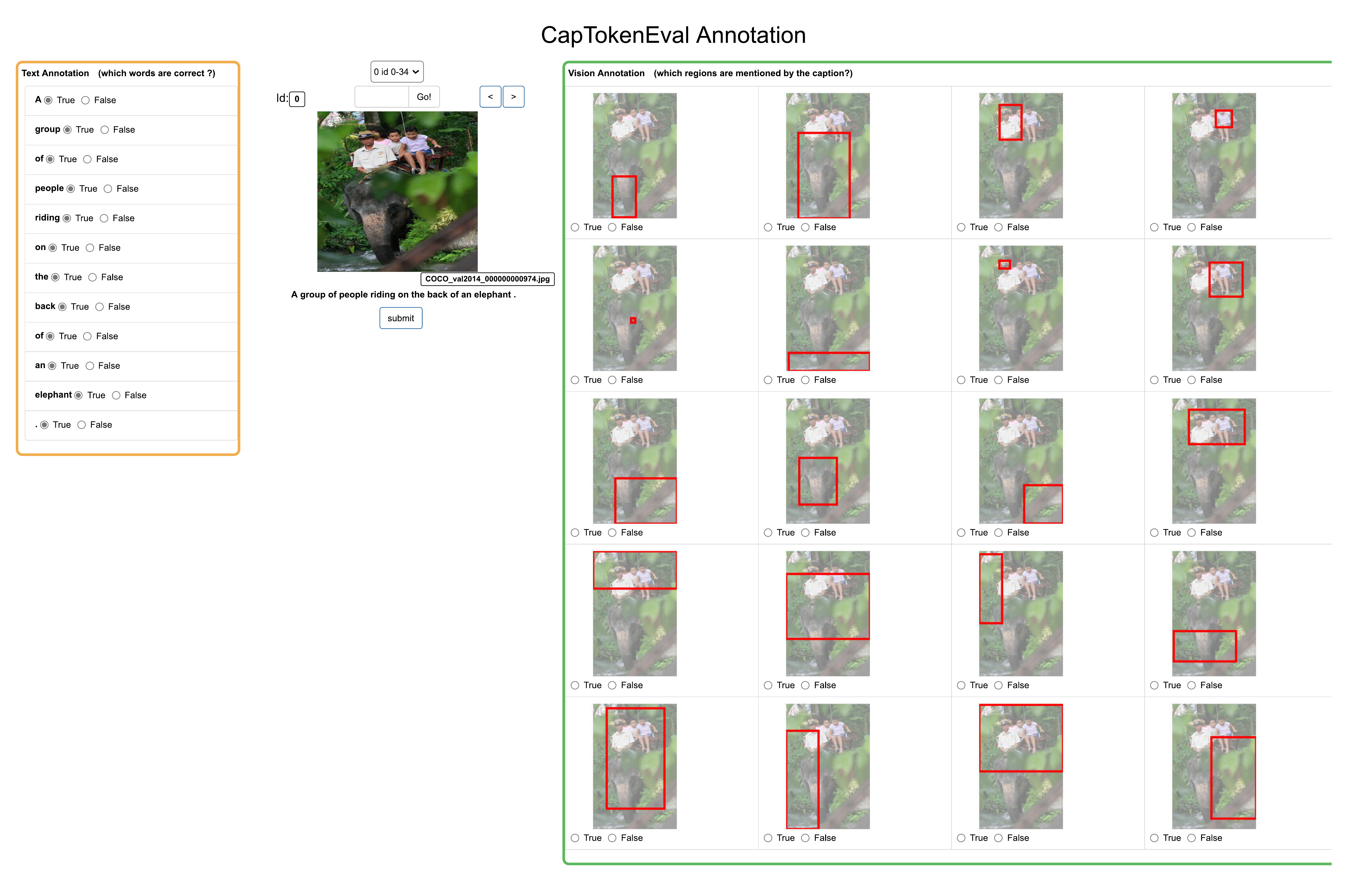}
    \caption{An overview of the annotation platform.}
    \label{fig:platform}
\end{figure*}

We build a platform to support the fine-grained annotation. Figure \ref{fig:platform} presents the annotation interface on our platform, which consists of three major parts. The middle part contains an image-caption pair to be annotated. The left part is the textual token-level annotation area, which lists all tokens in the caption. The right part is the visual token-level annotation area, which places 20 images with bounding boxes indicating different image regions.

\subsection{Annotation Instruction}
\label{instruction}
Given an image-caption pair, we ask annotators to identify which tokens in the caption are incorrect and which regions are mentioned by the caption. Besides, we require that if the caption mentions an object without descriptions about details, the image regions of detailed components shouldn't be classified as `Mentioned'. For example, for the caption `a group of people riding on the back of an elephant', the image region of the elephant nose shouldn't be judged as `Mentioned'.

We invite 20 college students as annotators. They all have sufficient English proficiency to understand image captions in English. We provide a document to inform annotators the goal of our annotation and detailed instructions about the usage of the annotation platform. Each annotator is assigned 35 image-caption pairs for annotation.

\end{document}